\documentclass[journal]{IEEEtran}

\ifCLASSINFOpdf
 
\else
 
\fi

\usepackage{amsmath}
\usepackage{multirow}
\usepackage{algorithmic}
\usepackage{graphicx} 
\usepackage{subfigure}
\usepackage{booktabs,lscape}
\usepackage{xcolor}
\usepackage[lined,boxed,commentsnumbered]{algorithm2e}
\usepackage{bm}

\usepackage{cite}
\usepackage[pagebackref=true,breaklinks=true,letterpaper=true,colorlinks,bookmarks=false,citecolor=blue,linkcolor=blue]{hyperref}

\begin{document}
 
\title{VPFNet: Improving 3D Object Detection with Virtual Point based LiDAR and Stereo Data Fusion}

 \author{Hanqi~Zhu, Jiajun~Deng, Yu~Zhang, Jianmin~Ji, Qiuyu~Mao, Houqiang~Li,~\IEEEmembership{Fellow,~IEEE},  Yanyong~Zhang,~\IEEEmembership{Fellow,~IEEE}

\thanks{Yanyong~Zhang is the corresponding author.
 
Hanqi~Zhu, Yu~Zhang, Jianmin~Ji, Qiuyu~Mao, Yanyong~Zhang are with School of Computer Science and Technology, University of Science and Technology of China, Hefei, China (e-mail: zhuhanqi@mail.ustc.edu.cn, yuzhang@ustc.edu.cn, jianmin@ustc.edu.cn,  qymao@mail.ustc.edu.cn,  yanyongz@ustc.edu.cn). 

Jiajun~Deng, Houqiang~Li are with Department of Electric Engineering and Information Science, University of Science and Technology of China, Hefei, China (e-mail:  dengjj@mail.ustc.edu.cn, lihq@ustc.edu.cn) }

}

\markboth{ }%
{Shell \MakeLowercase{\textit{et al.}}: Bare Demo of IEEEtran.cls for IEEE Journals}

\newcommand{\systemname}{{\sf VPFNet}\xspace}

\maketitle
 
\IEEEpeerreviewmaketitle

\begin{abstract}

It has been well recognized that fusing the complementary information from depth-aware LiDAR point clouds and semantic-rich stereo images would benefit 3D object detection. Nevertheless, it is not trivial to explore the inherently unnatural interaction between sparse 3D points and dense 2D pixels. To ease this difficulty, the recent proposals generally project the 3D points onto the 2D image plane to sample the image data and then aggregate the data at the points. However, this approach often suffers from the mismatch between the resolution of point clouds and RGB images, leading to sub-optimal performance. Specifically, taking the sparse points as the multi-modal data aggregation locations causes severe information loss for high-resolution images, which in turn undermines the effectiveness of multi-sensor fusion. In this paper, we present \systemname---a new architecture that cleverly aligns and aggregates the point cloud and image data at the `virtual'  points. Particularly, with their density lying between that of the 3D points and 2D pixels, the virtual points can nicely bridge the resolution gap between the two sensors, and thus preserve more information for processing. Moreover, we also investigate the data augmentation techniques that can be applied to both point clouds and RGB images, as the data augmentation has made non-negligible contribution towards 3D object detectors to date. We have conducted extensive experiments on KITTI dataset, and have observed good performance compared to the state-of-the-art methods. Remarkably, our \systemname achieves 83.21\% moderate 3D AP and 91.86\% moderate BEV AP on the KITTI test set, ranking the 1st since May 21th, 2021. The network design also takes computation efficiency into consideration -- we can achieve a FPS of 15 on a single NVIDIA RTX 2080Ti GPU.  The code will be made available for reproduction and further investigation.

\end{abstract}

\begin{IEEEkeywords}
3D Object Detection, Point Clouds, Stereo Images, Multiple Sensors

\end{IEEEkeywords}

\section{Introduction}

\IEEEPARstart{A}{ccurate} and timely 3D object detection is an important yet challenging problem in many real-world 3D vision applications, including autonomous vehicles, domestic robots, virtual reality, \emph{etc}. As explored in the literature, 3D object detectors can be developed combining data from different sensors, such as monocular images~\cite{pseudolidar}, stereo images~\cite{rts3d,plume} and LiDAR point clouds~\cite{3dssd-net,pointrcnn}. Although significant progress have been witnessed, the detection accuracy achieved by individually processing data of a single sensor is far from sufficient, due to the limitation of each sensor. For example, the RGB images are rich in semantic attributes, but lack depth information; the point clouds offer precise distance measurements, but cannot provide semantic information like colors, textures, and fine-grained shapes.

\begin{figure}[t]
\centering
\vspace{-8pt}
 
\includegraphics[width=\linewidth]{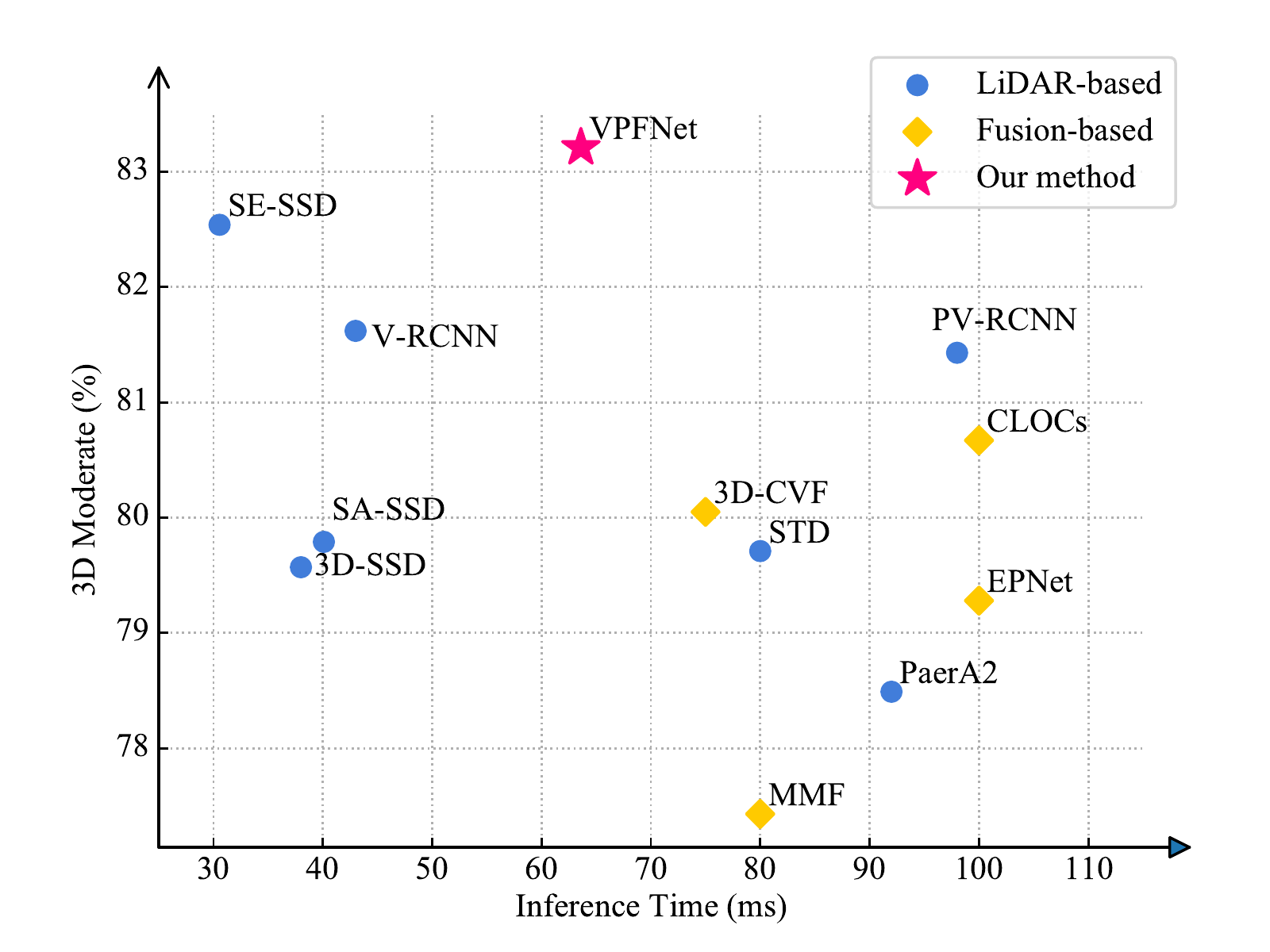}
 
\caption{Our proposed \systemname achieves the highest accuracy in 3D object detection for the moderate of car category on the KITTI~\cite{kitti-net} benchmark. Compared with the other multi-modal 3D object detection methods (marked in yellow), \systemname  achieves a competitive inference time of 63.6ms. Please refer to Table~\ref{accuracy} for a more comprehensive performance comparison. \label{fig:detection}}
\end{figure}

In light of the properties of RGB images and LiDAR point clouds, several innovations have been made to leverage their complementary information  
These multi-modal approaches  
can generally be categorized into two groups: object-level fusion~\cite{avod-net,mv3d-net,fconv-net,fpointnet,clocs-net,SCANET,SIFRNet} and point/voxel-level fusion ~\cite{3dcvf-net,epnet-net,mmf,painting,continues-net,pircnn,mvx-net} according to the fusion granularity.

The first group of schemes combines the features at the object or region of interest (RoI) level. For example, AVOD~\cite{avod-net}  combines  each  object’s intermediate RoI features  including  image  features  and  point  cloud  bird’s  eye view  (BEV)  features via simple feature concatenation. SIFRNet~\cite{SIFRNet} utilizes image features and frustum point clouds to generate 3D detection results following~\cite{fpointnet}. These schemes combine high-level, coarse-grained features, implicitly encoding the geometry information into the RoI context.

The second group of schemes conducts point/voxel level fusion, exploiting information fusion at a much finer granularity. For example, PointPainting~\cite{painting} projects LiDAR points to the image plane, and retrieves the classification information  from image semantic segmentation. EPNet~\cite{epnet-net} also projects points to the image plane,  retrieving image features instead of semantic masks.  3D-CVF~\cite{3dcvf-net} constructs a 3D volume, and maps image features to the BEV plane. Multi-Task Multi-Sensor Fusion (MMF)~\cite{mmf} projects the 2D depth maps to the 3D space and forms pseudo LiDAR point clouds, then performs continuous fusion~\cite{continues-net} at multiple granularities.

In this work, we take the point/voxel-level fusion approach in favor of its fine fusion granularity.  These schemes usually involve combining low-level multi-modal features at 3D points or 2D pixels. However, as pointed out in~\cite{review}, these schemes usually face a problem referred to as ``resolution mismatch''. That is, the mismatch between dense image pixels and sparse LiDAR points renders that only a small portion of pixels have matching points. 

\begin{figure}[!t]
\centering
\begin{tabular}{cc}
\includegraphics[width=0.5\columnwidth]{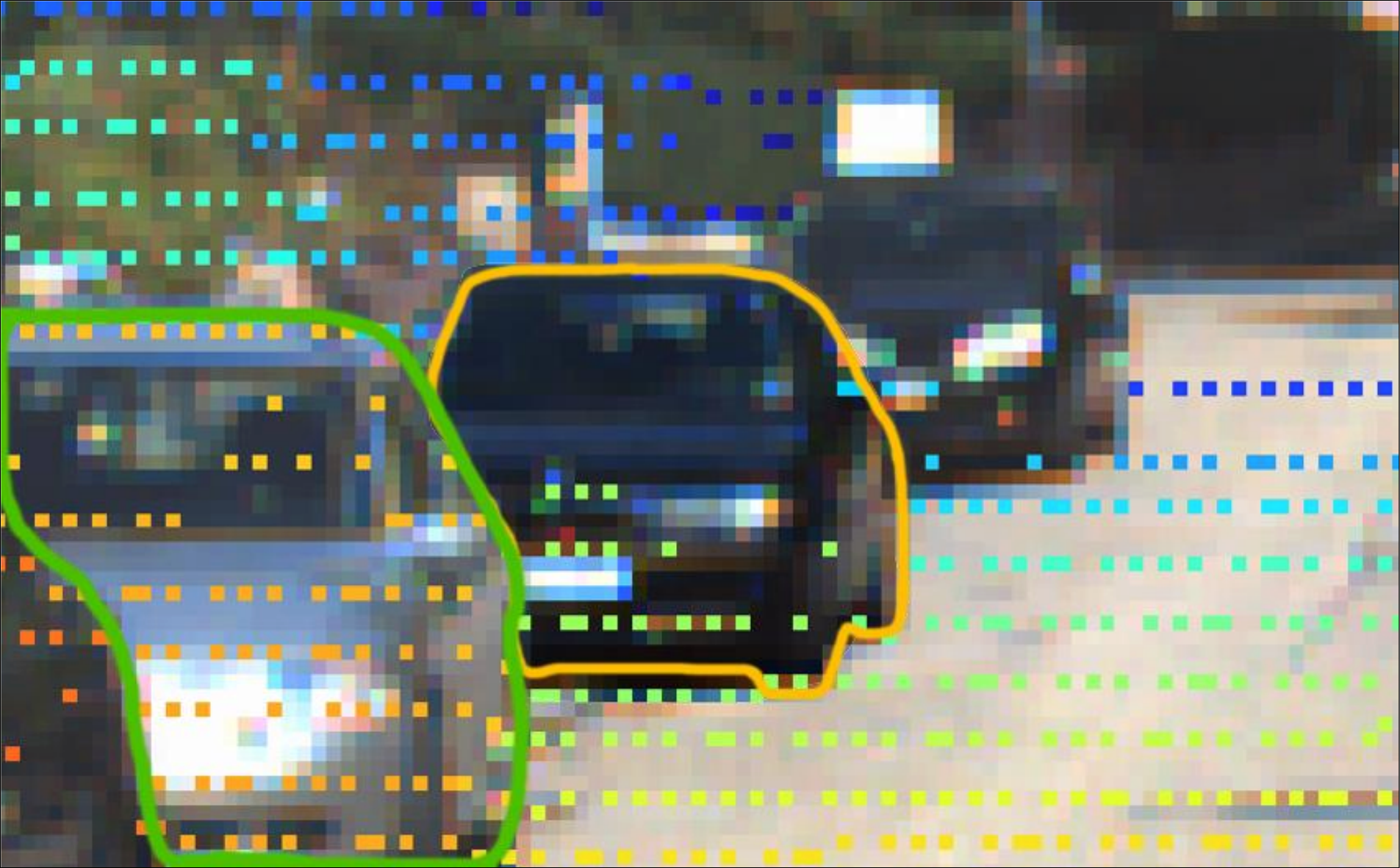}  \\
\end{tabular}
 \caption{\label{fig:mismatch} Illustration of resolution mismatch on the KITTI dataset. The car marked in orange contains 23 LiDAR points and more than 900 pixels. With the current sensor technology, pixel density is usually much higher than the point density. }
\end{figure}

We further illustrate the problem in Figure~\ref{fig:mismatch}. It shows a black car that is more likely to absorb the laser energy, causing resolution mismatch. In fact, on the KITTI dataset, a Velodyne-64 LiDAR point cloud has only 20,000 points on the camera view while an image has more than 300,000 pixels.  
If we directly project 3D point clouds to the 2D plane and sample the features at matching image pixels~\cite{painting}, we will have a rather low sampling rate for image data, roughly 6.7\%. Such a low sampling rate leads to serious information loss in the image plane. 

Here, as an alternative approach, we could try to lift 2D image features to the 3D space as in~\cite{3dcvf-net}. However, naively associating the 2D pixels with the entire 3D space, this approach will significantly increase the processing demand. In consideration of computation efficiency, these schemes usually need to down-sample the aggregated features, e.g., further projecting these features to the BEV plane for subsequent processing. In this case, we may lose valuable 3D information.

To address this dilemma, we cleverly employ a set of ``virtual'' points in the 3D space that act as the multi-modal data aggregation points, whose density is in between pixel density and point density, $D_{point} < D_{virtual} < D_{pixel}$. For example, in our implementation, the density of virtual aggregation points is roughly $20$ times the density of actual LiDAR points in foreground areas.  In this way, we can effectively bridge the resolution gap between the two sensors. When we project virtual points to the image plane, the resulting image data sampling rate will be higher. 

However, when we try to use virtual  points to aggregate corresponding LiDAR point data, a new challenge arises --  virtual points are not directly associated with any point cloud features. To solve this challenge, we leverage the voxel-RoI pooling operation discussed in~\cite{voxelrcnn} that can locate the $K$ nearest point neighbors in the 3D space for a given location.  By aggregating $K$ nearby point features to the virtual  point, we also effectively replicate the sparse point cloud features, essentially leading to a much denser point cloud features.

Using virtual points to aggregate multi-modal data for the first time, we refer to our method as \emph{virtual-point fusion}, \systemname in short. In addition to the novel virtual-point aggregation, we also explore several multi-modal data augmentation techniques. Multi-modal detectors often face the difficulty of applying data augmentation to both types of sensor data, such as global rotation augmentation and ground truth boxes sampling augmentation. We carefully extend these techniques to multi-modal cases.
 Our evaluation results on KITTI show that \systemname is ranked the highest for the following three test scenarios: $AP_{3D}$ moderate (83.21\%),  $AP_{BEV}$ moderate (91.86\%), and $AP_{BEV}$ hard (86.94\%). Our test performance on $AP_{3D}$ hard (78.20\%) is ranked the highest among all the published results. Further, compared with our point cloud only network Voxel-RCNN~\cite{voxelrcnn},  \systemname can improve the $AP_{3D}$  value by  $0.12\%$, $1.59\%$, and $1.14\%$ for easy, moderate, and hard on the \emph{car} category, respectively. In Figure~\ref{fig:detection}, we show the performance (both $AP_3D$ moderate accuracy and inference time) of \systemname and several recent 3D detection networks (including both fusion-based networks and LiDAR-based networks) -- \systemname has the most accurate detection among them all, and its inference latency is the lowest among all the fusion-based detectors.

To summarize, this paper has the following contributions:
 
\begin{itemize}

    \item  We devise an accurate two-stage voxel-level stereo-LiDAR fusion framework, \systemname, to effectively alleviate the resolution mismatch problem in fusing LiDAR and camera data. Easing this problem can nicely balance high sensor data sampling rate and high computation efficiency.

    \item  We explore several techniques to optimize our \systemname network, including random 3D proposal resizing to prevent over-fitting, weighted multi-branch combination to prevent single-modality dominance, and cut-n-paste based data augmentation.

    \item Over the popular KITTI dataset, we have carefully evaluated our 3D detection network.  The results show that \systemname   delivers the best performance in several test scenarios.  Moreover, with our light-weight design, the above accuracy can be achieved at a processing rate of 15 FPS.  
    
\end{itemize}

\section{Related Work}\label{section:related}

In this section, we present the background and related work for the following related topics: (1) LiDAR-only 3D object detection, (2) Stereo-only 3D object detection, (3) LiDAR and camera fusion based 3D object detection (4) Augmentation for 3D Object detection

\vspace{4pt}\noindent\textbf{3D Object Detection Using LiDAR Only:} As far as LiDAR-based 3D object detection is concerned, there are two mainstream models: point-based models and voxel-based models. Point-based methods~\cite{qiu2021geometric, chen2020hapgn,liu2020semantic} use point set abstraction to sample a fixed number of points as keypoints,  and aggregate point features around keypoints with ball query. The keypoints preserve the original location information. However,  since the point clouds are in-ordered, the complexity of point query  is O(N), where N denotes the number of point clouds.

Voxel-based models~\cite{voxelnet,second-net,Hollow-3D} transfer raw-points to voxels. Voxel representation keeps the memory locality~\cite{pvcnn}, while loses the original location information. SECOND~\cite{second-net} introduces a novel sparse convolution layer targeted at LiDAR point clouds. Voxel-RCNN ~\cite{voxelrcnn} devises voxel query  to take the place of ball query, and achieves the state-of-the-art accuracy and high efficiency on the KITTI benchmark. 

 The above studies consider how to effectively perform 3D object detection usinwg LiDAR point clouds. Our work shows that by considering both RGB images as well as point clouds, we can achieve better 3D detection performance.  

 \vspace{4pt}\noindent\textbf{3D Object Detection Using Stereo Only:}  
 There are several works that lift stereo images to the 3D space. They can be divided into two groups.  One is point based methods~\cite{CDN,pseudolidar++,confi3d}. These methods generate intermediate 3D structure of pseudo LiDAR point clouds by depth estimation, and feed the points into LiDAR-only 3D detectors. These methods generate huge amount of pseudo points and is time-consuming.

Another group  constructs cost volume in 3D space. For example, PLUME~\cite{plume}  constructs feature volume in 3D space and compresses it to BEV, then applies 2D convolution network to generate final results. The recently proposed two-stage method RTS3D~\cite{rts3d} constructs  instance level volumes using proposals, and uses an attention module and MLPs to get the final results. 

 The above studies consider how to effectively perform 3D object detection using stereo images in 3D space.  Our work shows that by considering both RGB images as well as point clouds, we can achieve better 3D detection performance. 
 
\vspace{4pt}\noindent\textbf{3D Object Detection Using Both LiDAR and Cameras:}   
Several multi-modal 3D object detection networks have been proposed~\cite{Survey}.  There are two levels of  fusion: object level fusion~\cite{avod-net,mv3d-net,fconv-net,fpointnet,clocs-net,SCANET,SIFRNet} and point/voxel level fusion~\cite{3dcvf-net,epnet-net,mmf,painting,continues-net,pircnn,mvx-net}. Object level fusion methods fuse in a coarse granularity. For example, AVOD~\cite{avod-net}  fuses features extracted from LiDAR BEV point clouds and camera front-view image and feed the features into fused RPNs for object detection; SIFRNet~\cite{SIFRNet}, F-ConvNet~\cite{fconv-net}  utilize front view images and frustum point clouds to generate 3D detection results following~\cite{fpointnet}; CLOCs~\cite{clocs-net} encodes an object's 2D and 3D detection results into a sparse tensor, and uses a 2D convolution network to refine the result.

The second group of schemes conduct point/voxel level fusion methods, they explicitly project the data between the 2D and 3D space, and fuse in a much finer granularity.
For example, PointPainting~\cite{painting} and EPNet~\cite{epnet-net} project LiDAR points to the image plane, and retrieve segmentation masks or image features. The point clouds are decorated with the retrieved features/masks. continuous-fusion~\cite{continues-net} associates image features and point cloud features by adopting KNN search, and uses the non-empty neighbour features to be the empty voxel feature. These methods exploit only the sparse information contained in a dense image, and suffer from information loss. The problem is referred to as resolution mismatch~\cite{review}.
 
 3D-CVF~\cite{3dcvf-net} constructs a high resolution 3D voxel space and lifts image features to the dense 3D voxel space via camera matrix. To avoid expensive 3D convolution, it compresses the height channel and converts 3D voxels to BEV feature map, and loses the 3D information. To explore fusion in the second stage, 3D-CVF divides the proposal into grid points and projects them to the image plane and retrieves 2D image features. However, the extracted 2D image features are directly aggregated by PointNet~\cite{pointnet} without any interaction with the 3D space.  
 
MMF~\cite{mmf} aims at multiple relevant tasks learning. It exploits depth estimation and converts depth maps into pseudo-LiDAR points. The computation overhead is heavy if pseudo-LiDAR points are generated at pixel-level. 

 Building upon the success of the above pixel-level fusion schemes, we further consider how to bridge the resolution mismatch of these two modalities by employing ``virtual'' points in the 3D space.  Our work shows that this step can lead to improved detection performance. 

\vspace{4pt}\noindent\textbf{Augmentation for 3D Object Detection:}  
 As proposed in~\cite{pv-rcnn}, popular point cloud augmentations include global rotation, global flipping, global scaling, and GT-sampling.  These augmentations can  accelerate convergence and boost the detection performance~\cite{second-net}. For global augmentations, they are reversible, and can be easily applied to fusion. However  GT-sampling is not designed for multi-modal fusion.  Some existing works try to avoid multi-modal GT-sampling  by training with only global augmentation~\cite{epnet-net}, or using GT-sampling only when the LiDAR backbone network was pretrained~\cite{3dcvf-net}.  Recently, a new cut-n-paste based  scheme was proposed in~\cite{augmenting}.  Inspired by~\cite{yun2019cutmix}, it cuts point cloud and image patches of ground-truth objects and pastes them into different scenes during training.  

 In addition to the above augmentations, we further employ a cut-n-paste based object sampling scheme that leverages the masks from prediction. Our work shows that this augmentation technique can effectively improve the performance.  
 
 \section{ Preliminaries on Point/Voxel-Level LiDAR-Camera Fusion  }
 
 Since \systemname performs sensor fusion at the point/voxel level,  we now provide a background overview on the architecture of this approach. A typical point/voxel level fusion scheme~\cite{painting,epnet-net,continues-net} consists of four steps: (1) 3D to 2D projection,  (2) 2D feature sampling, (3) multi-modal data aggregation, and (4) 3D  object detection.  
 
 Existing schemes often directly use LiDAR points as multi-modal data aggregation points~\cite{painting,epnet-net,mvx-net}. They project the LiDAR point $P$ to the image plane as follows:
 
\[
		z_c \left[
		\begin{matrix}
			u\\
			v\\
			1
		\end{matrix}
		\right]  = h K
		\left[
		\begin{matrix}
			R  & T 
		\end{matrix}
		\right]  
		\left[
		\begin{matrix}
			P_x\\
			P_y\\
			P_z \\
			1
		\end{matrix}
		\right],
\]
 where $P_x,P_y,P_z$ denote $P$'s 3D location, $u, v, z_c$ denote the 2D location and the depth of its projection on the image plane, $K$ denotes the camera intrinsic parameter,  $R$ and $T$ denote the rotation and the translation of the lidar with respect to the stereo camera reference system. and $h$ denotes the scale factor due to down-sampling.

\begin{figure}[!t]
\centering
\includegraphics[width=3.5in] {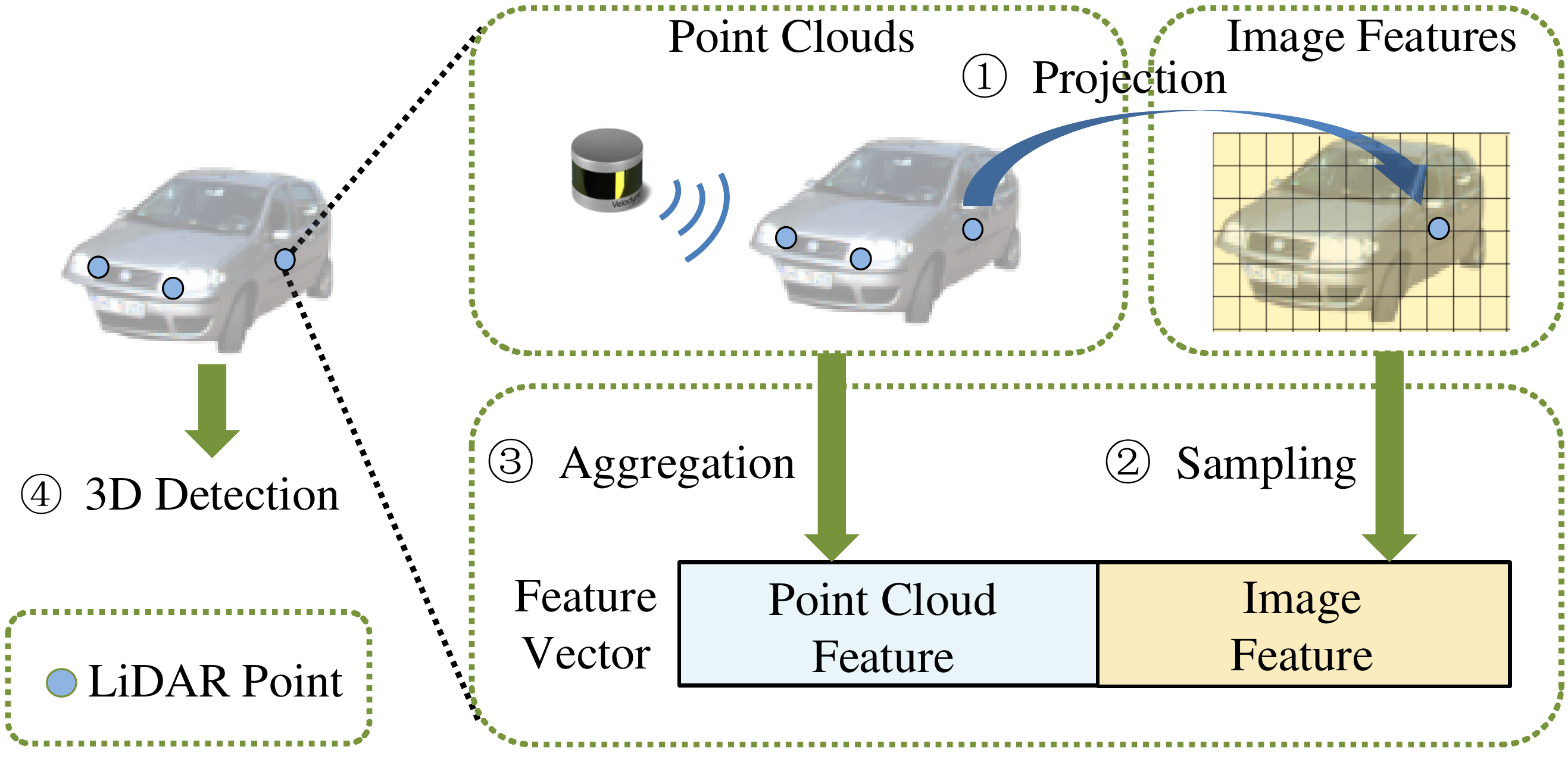} 
\caption{In a point/voxel level fusion scheme, the original or downsampled LiDAR points are projected to the 2D plane, to retrieve the corresponding 2D image features. The multi-modal features are then aggregated by concatenation for subsequent 3D object detection. }
 
\label{fig:framework}
\end{figure}

 Then the corresponding image features at location $(u, v)$, $F_{u, v}$, can be sampled with bilinear interpolation. The sampled feature is then concatenated with point $P$'s own location and feature to form an aggregated \emph{feature vector}. Figure~\ref{fig:framework} illustrates this process. 
 
 For example, in PointPainting~\cite{painting}, it aggregates the feature vector with a simple concatenation operation as follows: 
      \[ \bm{\phi}_{P}  =  \text{concat}\{P_x;P_y;P_z;P_i;mask_{(u,v)}\}, \]
     where $\bm{\phi}_{P}$ is the aggregated feature vector at LiDAR point $P$, $P_{x}$, $P_{y}$, $P_{z}$, $P_{i}$ are the 3D coordinates, and intensity of point $P$, $mask_{(u,v)}$ is the sampled image segmentation mask.   
As another example, EPNet~\cite{epnet-net} utilizes a L1-fusion block to aggregate the multi-modal data in a feature vector. 
 
Finally, the aggregated feature vectors are fed to the detection head. The final box classification  and  regression results are generated accordingly.

\section{ Virtual-Point Fusion (~\systemname)  } \label{design}

  In this section we present the design of \systemname.  We first give the design overview, and then describe  the important components one by one.

\subsection{Design Overview}  

\begin{figure*}[!t]
\centering
\includegraphics[width=7in] {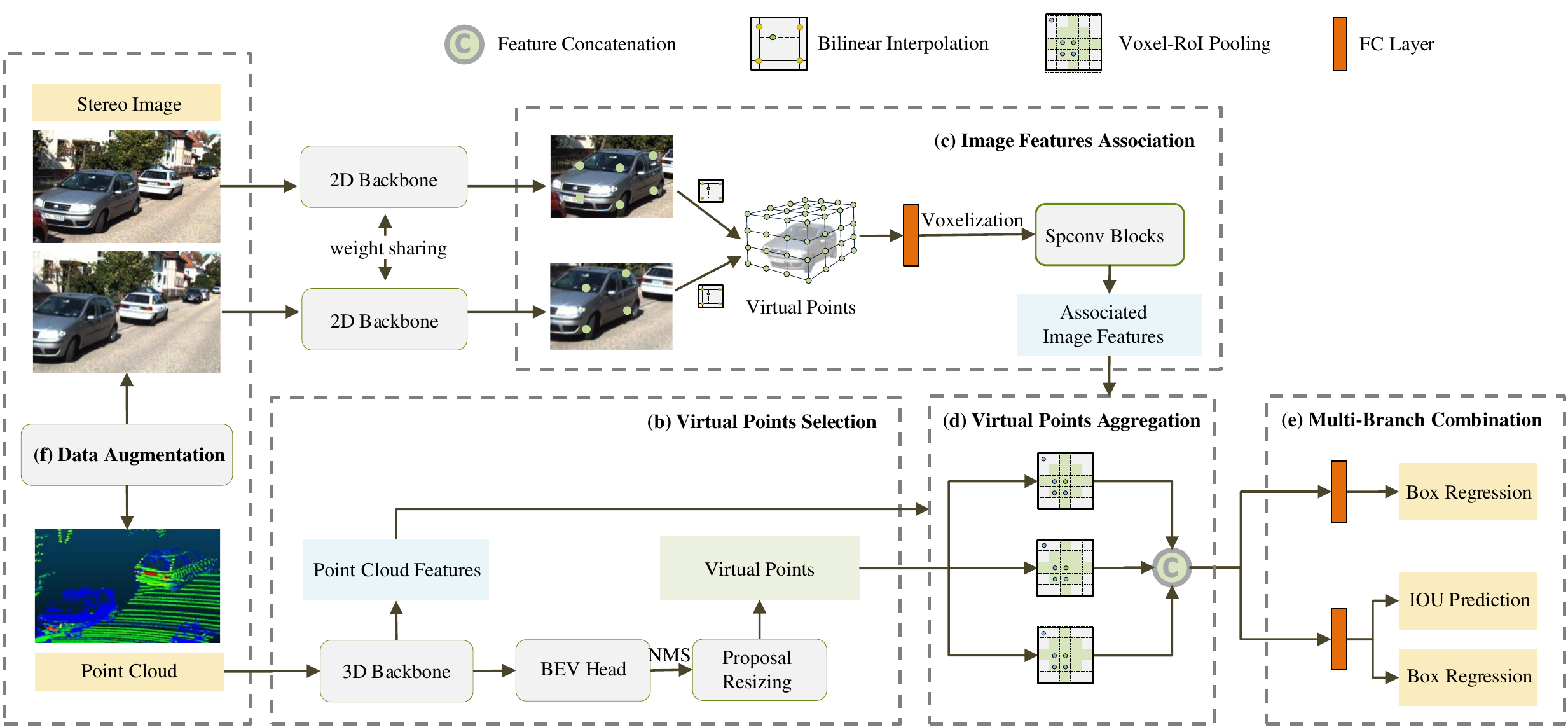} 
 
\caption{   \label{fig_DRF}The architecture of the proposed \systemname. Each dashed line box is associated with  a subsection in Sec~\ref{design}. Given a stereo and LiDAR point cloud pair, we first use (f) multi-modal data augmentation techniques to obtain an augmented scene in the training stage. Then in (b) virtual point selection, we feed the data into the 2D and 3D backbones separately. The 2D backbone is used only for feature extraction, and the 3D backbone is for generating the 3D proposals.  The 3D proposals are randomly resized and divided into fixed-sized grid blocks, and the resulting vertices of each grid block/cell are virtual points.  Then in (c) image feature association, the virtual points are projected to the image plane to sample the 2D image features. The sampled image features are thus associated with the virtual point. These features are voxelized and processed by stacked sparse convolution layers. Next, in (d) virtual points aggregation,we sample the virtual points as query points, and locate the $K$ nearest LiDAR point neighbors in the 3D space, and aggregates their features with the associated images, resulting in an aggregated feature vector. Finally  In (e) multi-branch combination, these feature vectors are fed into the detection head to predict the final results. We also involve an auxiliary branch that only processes the aggregated image features to prevent single-modality dominance.  Please note that the augmentation step and the proposal resizing step are only needed in the training stage. }

\end{figure*}

We illustrate our multi-modal 3D detection network in Figure~\ref{fig_DRF}. 
The primary principle in designing \systemname is to (1) carefully \emph{bridge the resolution gap} between LiDAR point clouds and camera images, and (2) balance the sampling rate needed for detection accuracy and computation efficiency.  

The first design choice is to determine whether we project 3D data to 2D to perform fusion, or vice versa. Considering the properties of images and point clouds (summarized in Table~\ref{modality}), we choose to aggregate the sensor data in the 3D space. In this way, we can involve  most of the point cloud information in the processing and avoid scale effect. Scale effect means an object's 2D area size is related with both its actual sizes and depth. In general, LiDAR points are much more sparse than image pixels and contain less redundancy. They also give more accurate 3D detection results than images. As such, we argue that a fusion network needs to keep as much point cloud information as possible. 

The second design choice is to determine at what locations in the 3D space to aggregate the data from both sensors.  In order to bridge the resolution gap between the two, we choose to do so at \emph{virtual} points that are sparser than image pixels but denser than LiDAR points.  Given a virtual point, we need to find the corresponding image features and point cloud features, and suitably aggregate them. For image features association, we project the virtual point's location to the 2D image plane, and retrieve the left/right image features accordingly.  For point cloud  features association, we perform a voxel-RoI pooling operation~\cite{voxelrcnn}  to locate neighbors and aggregate their features. In this way, we sample the dense image features while aggregating/replicating the sparse point cloud features, nicely balancing the disparity in their resolutions.

    \begin{table}[b]
    \centering
    
    \begin{tabular}{lll} 
    \hline\noalign{\smallskip}
     & Point Cloud & Image  \\
    \noalign{\smallskip}\hline\noalign{\smallskip}
    Feature Aggregation & 3D neighbours & 2D neighbours\\
    Permutation &  invariant~\cite{pointnet} & variant \\
    Scale Effect & no  & yes \\
    Memory Access &  unordered~\cite{pvcnn}   & ordered \\
    Resolution & sparse & dense \\
    \noalign{\smallskip}\hline
    \end{tabular}
    ~\vspace{5pt}
    \caption{\label{modality}  Comparison of point clouds and images}
 
    \end{table}

\begin{figure}[!t]
\centering
\includegraphics[width=3.5in] {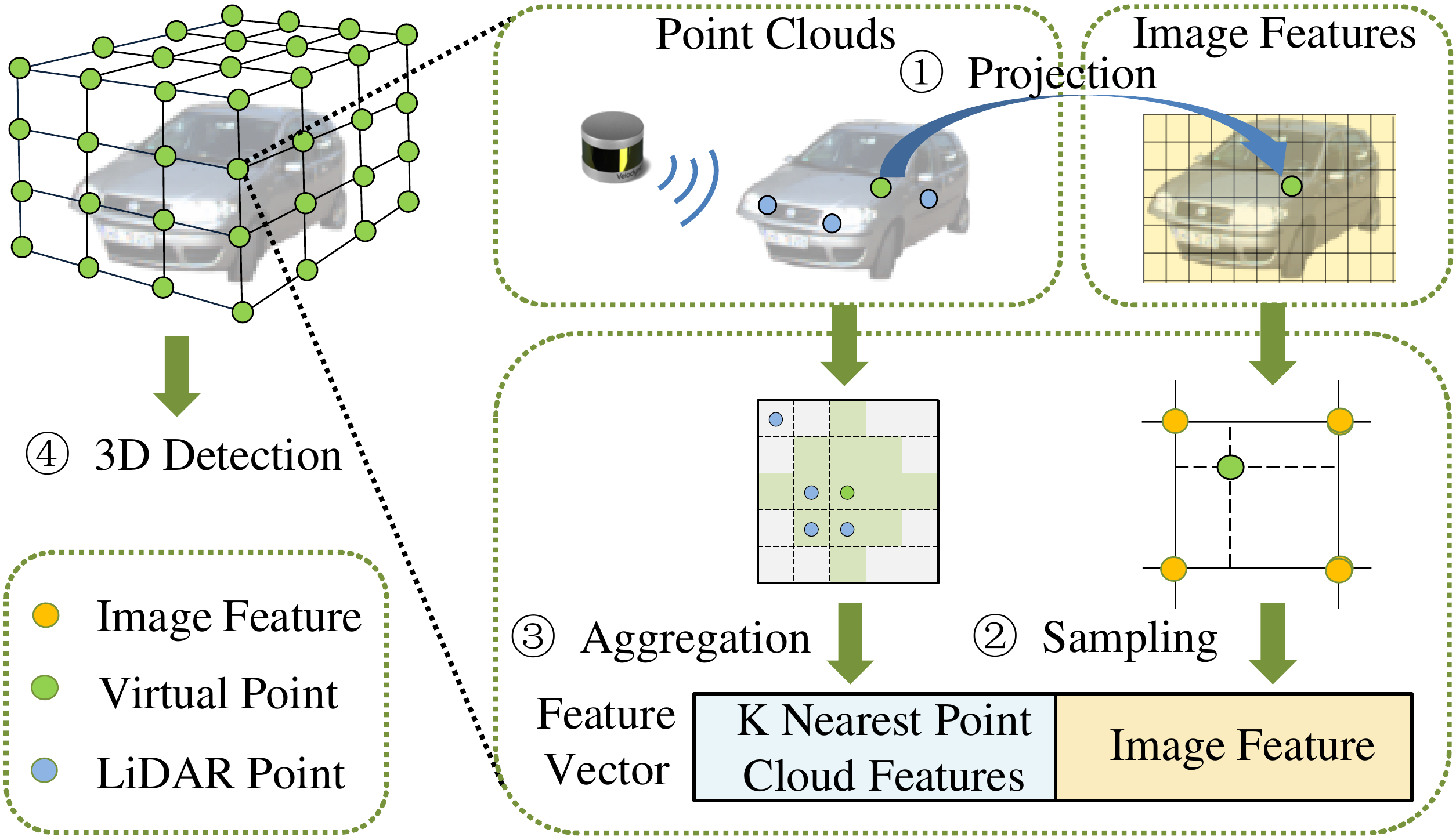} 
\caption{  \label{fig_DRF_fv}In \systemname,  we use virtual points (marked in green) to aggregate the multi-modal data. Virtual points are projected to the image plane to sample image features. As far as LiDAR point cloud features, we perform a voxel-RoI pooling operation (marked by the $5\times5$ square) to aggregate point cloud features from $K$ nearest neighbors. The feature vector at each virtual  point thus consists of the sampled image feature and aggregated $K$ point cloud features.}
\end{figure}

Specifically, as shown in Figure~\ref{fig_DRF}, our network has the following main components:

\begin{enumerate}

     \item \emph{Virtual Points Selection}. This step selects the  virtual  points from 3D proposals. Specifically, we feed point clouds into 3D backbone networks and generate 3D proposals, randomly resize the 3D proposals for better robustness, and divide them into discrete grid blocks. We then use the resulting grid points as virtual  points (see Figure~\ref{fig_DRF_fv}).   
     
    \item \vspace{3pt} \emph{Associating Image Features to Virtual Points}. 
    This step samples the image features and associates the sampled image features to virtual points.  Firstly we generate 2D image features with a light-weight 2D backbone. Then we sample the image features through virtual point projection. We voxelize and convolve the sampled image feature vectors through stacked sparse convolution layers to extract more compact features.  
    \item \vspace{3pt} \emph{Aggregating LiDAR Point Features at Virtual Points}. This step aggregates suitable LiDAR features to each virtual point. We first use a modified voxel-RoI pooling layer to query and aggregate the sparse LiDAR point cloud features. Both LiDAR point cloud features and image features are aggregated in a multi-scale manner.  
 
    \item \vspace{3pt} \emph{Weighted Multi-Branch Combination}. This step predicts the final results. Here we consider the main branch as well as an auxiliary branch to avoid single-modality dominance and alleviate over-fitting. Both branches predict the final results, while supervised with different weights.
    
    \item \vspace{3pt} \emph{Data Augmentation}. This step performs suitable multi-modal data augmentation. 
\end{enumerate}
 
\subsection{Virtual Points Selection}
    The sample density of LiDAR point clouds is much lower than that of camera images, especially for faraway objects and black objects (see Figure~\ref{fig:mismatch}).  
    Due to the severe resolution imbalance, traditional fusion schemes often lead to low sampling rate for image features, which likely hurts the overall detection performance.
   
    To alleviate this problem, we do not use actual LiDAR points as multi-modal feature aggregation points, but propose to leverage \emph{virtual}  points (we sometimes refer to them as virtual points).  The density of such virtual  points is in between that of actual LiDAR points and image pixels. By attaching the suitable image features and point features to these virtual  points, we can effectively increase the image feature sampling rate, better match the two types of data,  and utilize more multi-modal features  in the 3D detection. 
    
    Next, we explain how we select virtual points in the 3D space. Firstly, we feed LiDAR point clouds into the 3D backbone network to generate 3D proposals and 3D feature maps.  Backbone refers to the network which takes the raw data as input and extracts the feature map upon which the rest of the network is based.   3D proposals refer to  3D candidate bounding boxes of the final detection  that need  to be verified in the refinement stage.   The 3D backbone network we use has the same structure as Voxel-RCNN in~\cite{voxelrcnn}. The 3D proposals go through a  non-maximum suppression (NMS) stage, while being ranked according to their intersection over union (IoU) with the ground truth. Only those proposals that have an IoU larger than a preset threshold will be selected. The proposals are further expanded by 0.8m  in all dimensions to cover more contextual information. 
    
    The virtual points are selected from within these proposals.  Since the proposals from point clouds are usually rather accurate, directly using them could potentially lead to over-fitting.  Therefore,  we  randomly resize the 3D proposals to boost the overall robustness of the model.
 
    Specifically, for each dimension of a 3D proposal, \systemname randomly picks a noise value from a uniform distribution and adds that noise value  to the original coordinate following the formulation below:
    \[\Delta k = \phi\{u_k\} , k \in \{x,y,z,w,h,l,\theta\}, \]
    where $\Delta k$ denotes the noise value in each dimension, and $\phi$ is a uniform distribution from $-u_k$ to $u_k$ ($u_k$ is varied according to different $k$), and $k$ denotes the seven dimensions of a box, including its box center location  x, y, z, box dimensions width, length, height, and $\theta$.   $\theta$ denotes  the orientation  angle of each box from the bird’s eye view. For each dimension $u_k$ is different. In Section~\ref{Experiment} we show that this proposal box resizing augmentation is rather effective.
    
    After random resizing, the proposals are divided to $N_x \times N_y \times N_z$ 3D grids along the length, width and height dimensions similar to~\cite{voxelrcnn}. We then use these grid points as the virtual points. Here, example $N_x$, $N_y$, and $N_z$ values are 12, 8, and 22,  resulting in more than 2K samples in a foreground area. Such a sample density is much higher than that of the original LiDAR point clouds (typically hundreds of samples).

\subsection{Associating Image Features to Virtual Points}

 	Given a virtual point, we first locate the corresponding image features and associate them to the virtual point.

\begin{figure}
\centering

\subfigure[2D backbone]{
\includegraphics[width=0.31\linewidth]{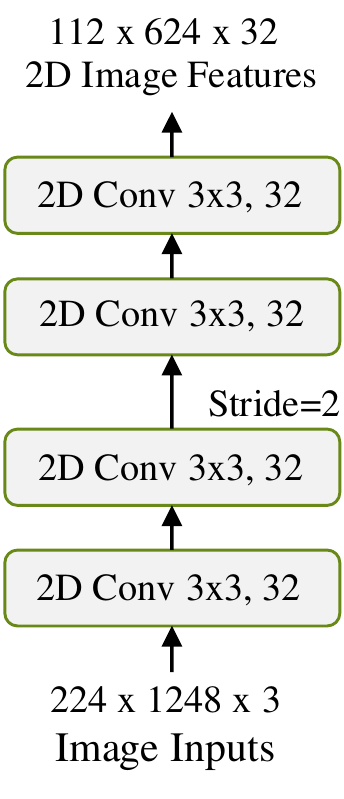}
}%
\subfigure[3D spconv block]{
\includegraphics[width=0.55\linewidth]{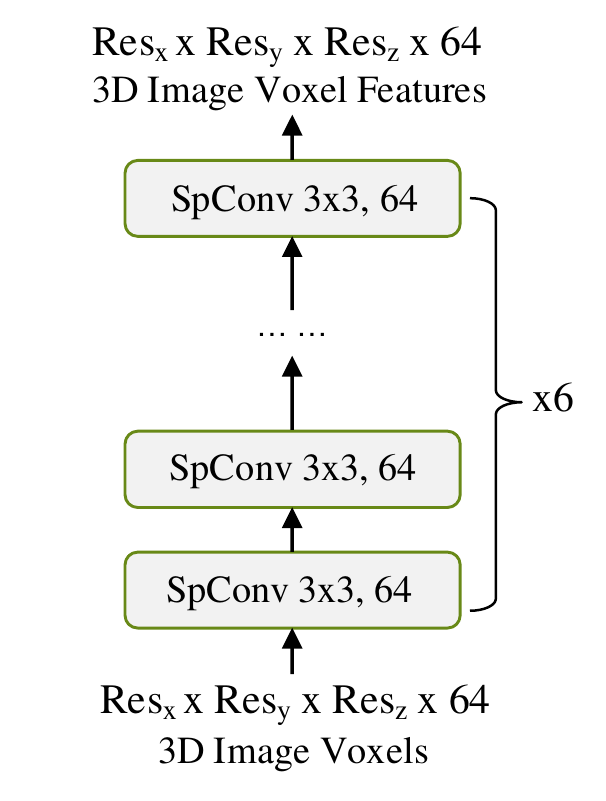}
}%

\centering
\caption{\label{fig:spconv}  The architecture of our 2D backbone (a) and 3D sparse convolution block (b). The output of the 2D backbone is down-sampled by two times  compared with its input. }
\end{figure}

	For 2D image feature extraction, we use two light-weight 2D convolution backbone networks. Following~\cite{rts3d,dsgn}, our 2D backbones are weight sharing. Figure~\ref{fig:spconv}(a) illustrates our 2D backbone structure. Our 2D backbone consists of four convolution layers, with all the convolution layers having a kernel size of 3$\times$3,  and the second layer having a stride of 2. The generated feature maps are thus 2$\times$ down-sampled compared to the original resolution.
 
We then attach the sampled left and right image features to the virtual  point $V$:
 \[ \bm{\phi_V}  =  \text{concat}\{ I_l;I_r;V_x;V_y;V_z\}, \]
 where $\bm{\phi}_{V}$ is the resulting feature vector after aggregating the image features associated with $V$ located at $(V_x, V_y, V_z)$, $I_l$ is the left image feature, $I_r$ is the right image feature. 

After aggregating the image features to the feature vector $\bm{\phi_V}$ at the virtual point, \systemname next voxelizes the feature vectors at all virtual points and performs feature extraction. In consideration of both information loss and computation efficiency, we choose to employ a light weight module that consists of six sparse convolution blocks~\cite{second-net} for feature extraction. Figure~\ref{fig:spconv}(b) illustrates our sparse convolution structure. All six spconv layers are  standard 3$\times$3 convolution layers. We obtain the following image feature map $D_0$:  
\[ D_0 =  SPCONV(\Phi), \]
where $\Phi$ is the universe set of all aggregated feature vectors, $SPCONV$ is the employed sparse convolution operation.

\subsection{Aggregating LiDAR Point Features to Virtual Points} 

Now we aggregate corresponding LiDAR point features to the virtual point. To compensate for the low sample density of LiDAR points, we locate \emph{multiple} nearby LiDAR points and aggregate their features at a virtual point. In this way, the LiDAR points' features are suitably replicated, effectively boosting the point cloud feature density. In order to find the nearby $k$ LiDAR points, we adopt the voxel-RoI pooling operation proposed in~\cite{voxelrcnn}.  To reduce the complexity, the query points here has a resolution of $G_x \times G_y \times G_z $, where  $G_x \times G_y \times G_z $ is the number of grid points within a 3D proposal. It is smaller than the resolution in last subsection. 
 
We next explain how we modify voxel-RoI pooling to fit our problem. 
Given a virtual point, we query and aggregate its feature neighbours on the feature map. Considering that the feature maps in \systemname are multi-modal and multi-scale, we execute queries on each feature map separately to pool the neighbouring features. The feature maps we use here in \systemname include the image feature map $D_0$, and LiDAR point cloud feature maps $D_1$ and $D_2$,  with $D_1$  being the 4x down-sampled feature map from our 3D backbone and $D_2$ the 8x down-sampled feature map.  The queried neighbours are then aggregated by accelerated PointNet~\cite{voxelrcnn}. The aggregated features are of the same shape, and can be easily concatenated. Accordingly, we update the aggregated feature vector as:
  \[ \bm{ \phi_{V} }  = \text{concat}\{Q^k( D_i )\}, \]
 where $D_{i}$ denotes the feature maps (3 feature maps in total as explained above),  $Q^{k}$ denotes the query operated on different down-sampling scales (2 scales in total, detailed in Section~\ref{Experiment}).  $\text{concat}{\cdot}$ denotes feature concatenation. For a voxel query $Q$, given a virtual point $v$, and a voxel feature map $\bm D$, it groups a set of $K$ neighbors $\{\bm{d}^1, \bm{d}^2,\cdots,\bm{d}^K\}$.  Then, all the $K$ neighbors are aggregated as follows:
    \[ Q(D) =  \max \limits_{k=1,2,\cdots,K}\{\Psi_1(\bm{d}^k-\bm{v}) +  \Psi_2(\bm F^k) \}, \]
    where $\bm{d}^k$ and $\bm{F}^k$ denote the location and voxel feature of the $k$-th  voxel neighbour, $\bm{v}$ denotes the location of the  virtual point $v$,  and $\Psi_1(\cdot)$ and $\Psi_2(\cdot)$ denotes two Multilayer Perceptron (MLP) layers as in~\cite{voxelrcnn}. A max pooling is then used to obtain the aggregated feature vector of the given virtual point as in~\cite{pointnet}.  
    
    Eventually, we have the aggregated feature vector $ \bm{ \phi_{V} }$ of dimensions $G_x \times G_y \times G_z \times 3C$,  $3C$ is the sum of  channel numbers of all three input feature maps. 
    
\subsection{Weighted Multi-Branch Combination}
    In this step, we generate the final detection result from all the aggregated feature vectors with detection heads. The detection head takes the virtual points as input and  predicts the box regression and the IoU prediction results.   In the detection head, the  aggregated feature vector of $G_x \times G_y \times G_z  \times 3C$ dimensions  is flattened and reduced to a $1 \times  512$ vector by MLPs as   in~\cite{pv-rcnn,voxelrcnn}. Since the virtual  points are sampled in a fixed order, we do not need to make the input permutation invariant. Hence, MLPs are sufficient to generate an accurate detection result.

     As observed in~\cite{multitower}, during training, a multi-modal network is likely dominated by one of the modalities. In our case, since the image features contain less 3D information than the LiDAR point features, LiDAR point features tend to dominate and lead to over-fitting. To address this issue, we propose to use an auxiliary detection branch to prevent over-fitting. 
    
     The auxiliary branch is similar to the main branch, with the difference that it takes the aggregated image features $D_0$ as input and  predicts only the box regression results. In this way, it enforces the image features to encode the 3D context to certain extent.  The two heads are assigned with different weights, and the loss function is  as follows: 
     \[
    L_{{rcnn}} =   L_{ {IoU}}  +   W_{ {V}}  \cdot L_{ {V}} +  W_{ {A }}  \cdot  L_{ {A}},\]
    where  $L_{{rcnn}}$ is the loss in the second stage, $ L_{ {IoU}}$ is the cross entropy loss for the 3D IoU prediction task in the main detection branch that regresses the 3D IoU between the predicted boxes and ground-truth boxes as in~\cite{voxelrcnn},  
    $ L_{ {V}} $ and  $ L_{ {A}} $  are the smooth-L1 loss for the 3D bounding box regression task in the main branch and auxiliary branch respectively that regresses the residual value of the predicted 3D box and the ground-truth, and  $ W_{ {V}} $ and  $ W_{ {A}} $ denote the loss weights for the two branches.
    
\subsection{Data Augmentation}
 
As a recent paper~\cite{augmenting} suggests, it is important to adopt multi-modal data augmentation techniques when multiple sensor types are used. As such, we devise  a  cut-n-paste based multi-modal data augmentation scheme.   To conduct cut-n-paste data augmentation, we firstly generate the foreground object mask list using a pre-trained instance level segmentation network, and generate the foreground object masks for images.

Additionally, we have also devised  a technique to help process stereo images. 
Here, we define a cost matrix and use the Hungarian algorithm~\cite{hungarian} to associate the predicted masks in the image pair. The score for each object pair is defined below:
   
    	\[\bm{Score}_{ij} =  \alpha  \cdot  \frac{\|v_i - v_j \|}{ h_i}   + \beta  \cdot \|s_i - s_j\|,\] 
   
    where $\bm{Score}_{ij}$ denotes the cost of left-right image pair, $v_i$ and $v_j$ denote the vertical location of the $i$-$th$ object in the left image and $j$-$th$ object in the right image, $h_i$ denotes the height of the 2D bounding box for the $i$-$th$ object,  $s_i$ and $s_j$  denote the object scores, and $\alpha$ and  $\beta$  denote the weight. 
    
   After we associate the objects in stereo images, we define a similar cost matrix to associate the stereo pair and the ground truth:
    	\[\bm{Score}_{kv} =   \frac{ \|v_k - v_v \| }{ h_k}   +    \frac{ \|u_k - u_v\|}{ w_k },\] 
    where $\bm{Score}_{kv}$ denotes the cost of stereo ground truth image pairs, $k$ denotes the $k$-$th$ stereo pair,  $v$ denotes the $v$-$th$ ground truth image, $h_k$ and  $w_k$  denote the predicted 2D bounding box height and width, and $v_k$ and $u_k$  denote the vertical and horizontal location of the predicted 2D bounding box center.

     After obtaining the matched stereo-GT triplets,  we then select the scenes from the training data according to their calibration files. Only those scenes that have exactly the same calibration information as the current scene will be selected. We randomly sample 100 objects from the selected scenes. Then we move all the objects' point clouds to the ground following~\cite{pv-rcnn,second-net}, together with the masks for consistency.  Then we sample 15 objects one by one and calculate 2D and 3D occlusion indicator. If it is less than our preset threshold, the object will be added to the current scene. Here, the occlusion indicator for the current object $I_i$ is calculated as:
 
    \[I_i=\max \limits_{k=1,2,\cdots,N}\{IoU_{i,k}\}, \]
    where $IoU_{i,k}$ is the IoU (3D and 2D) between object $i$ and the $k$-$th$ object in the current scene.   We consider the 3D and 2D occlusions separately. By adopting  occlusion threshold $\tau_{2D}$ and $\tau_{3D}$, we can avoid the unnecessarily challenging occlusion scenes caused by the augmentation process.  Finally we rearrange all the sampled objects according to their depth in ascending order to simulate occlusion, and paste both sampled point clouds and sampled object masks to the new scene.

\section{Evaluation} \label{Experiment}

\begin{table*}[t]
   \centering
   \footnotesize
   \begin{tabular}{|c|c|c|c|ccc|ccc| }
       \hline
       &
       \multicolumn{1}{c|}{ \multirow{2}{*}{Method}} & \multicolumn{1}{c|}{ \multirow{2}{*}{Venue}}& \multicolumn{1}{c|}{ \multirow{2}{*}{Sensor(s)}} & \multicolumn{3}{|c|}{$3D$} & \multicolumn{3}{|c|}{$BEV$}    \\
       \cline{5-10}
       &
       \multicolumn{1}{c|}{} & \multicolumn{1}{c|}{} & \multicolumn{1}{c|}{}& \multicolumn{1}{|c}{Easy} & \multicolumn{1}{c}{Mod} & \multicolumn{1}{c}{Hard} &
       \multicolumn{1}{|c}{Easy} & \multicolumn{1}{c}{Mod} & \multicolumn{1}{c|}{Hard}    \\
       \hline
       \hline
         \parbox[t]{2mm}{\multirow{13}{*}{\rotatebox[origin=c]{90}{One-stage}}}
    
         & VoxelNet~\cite{voxelnet} &CVPR 2018   &{LiDAR}      & 77.82  & 64.17  & 57.51     &87.95 &78.39 &71.29  \\
         & ContFuse~\cite{continues-net}         &ECCV 2018   &{LiDAR+RGB}    & 83.68  & 68.78  & 61.67     &94.07 &85.35 &75.88   \\
         & SECOND~\cite{second-net}      &Sensors 2018   &{LiDAR}      & 83.34  & 72.55  & 65.82       &89.39 &83.77 &78.59  \\
         & PointPillars~\cite{pointpillar}&CVPR 2019   &{LiDAR}      & 82.58  & 74.31  & 68.99     &90.07 &86.56 &82.81 \\
         & TANet~\cite{tanet-net}        &AAAI 2020   &{LiDAR}      & 84.39  & 75.94  & 68.82      &91.58  &86.54  &81.19   \\
         & Associate-3Ddet~\cite{associate}&CVPR 2020   &{LiDAR}      & 85.99  & 77.40  & 70.53      &91.40  &88.09  &82.96   \\
         & HotSpotNet~\cite{hotspot}  &ECCV 2020   &{LiDAR}      &87.60 &78.31 &73.34      &94.06 &88.09 &83.24   \\
         & Point-GNN~\cite{pointgnn}    &CVPR 2020   &{LiDAR}      & 88.33  & 79.47  & 72.29     &93.11  &89.17  &83.90   \\
         & 3DSSD~\cite{3dssd-net}       &CVPR 2020   &{LiDAR}      & 88.36  & 79.57  & 74.55   &92.66  &89.02  &85.86   \\
         & SA-SSD~\cite{sassd-net}    &CVPR 2020    &{LiDAR}      & 88.75  & 79.79  & 74.16     &95.03 &91.03 &85.96   \\
         & CIA-SSD~\cite{cia-ssd}    &AAAI 2021    &{LiDAR}      & 89.59   & 80.28   & 72.87      &93.74 &89.84 &82.39  \\

         & SE-SSD~\cite{se-ssd}    &CVPR 2021     &{LiDAR}      &  \bf91.49  &82.54  &77.15     &\bf95.68 &91.84 &86.72   \\

      \hline
      \hline
         \parbox[t]{2mm}{\multirow{13}{*}{\rotatebox[origin=c]{90}{Two-stage}}}
        & MV3D~\cite{mv3d-net}     &CVPR 2017   & {LiDAR+RGB}   & 74.97  & 63.63  & 54.00      &86.62 &78.93 &69.80   \\
         & F-PointNet~\cite{fpointnet} &CVPR 2018   & {LiDAR+RGB}   & 82.19  & 69.79  & 60.59       &91.17 &84.67 &74.77  \\
         & AVOD~\cite{avod-net}            &IROS 2018   & {LiDAR+RGB}   & 83.07  & 71.76  & 65.73      &89.75 &84.95 &78.32   \\
 
         & PointRCNN~\cite{pointrcnn} &CVPR 2019   & {LiDAR}     & 86.96  & 75.64  & 70.70       &92.13 &87.39 &82.72  \\
         & F-ConvNet~\cite{fconv-net}  &IROS 2019   & {LiDAR+RGB}   & 87.36  & 76.39  & 66.69       &91.51 &85.84 &76.11  \\
         & 3D IoU Loss~\cite{3dIoU}    &3DV 2019   & {LiDAR}     & 86.16  & 76.50  & 71.39       &91.36  &86.22  &81.20  \\
 
         & Fast PointRCNN~\cite{fast-point} &ICCV 2019   & {LiDAR}     & 85.29  & 77.40  & 70.24       &90.87 &87.84 &80.52  \\
         & UberATG-MMF~\cite{mmf} &CVPR 2019   & {LiDAR+RGB}   & 88.40  & 77.43  & 70.22      &93.67 &88.21 &81.99   \\
         & Part-$A^2$~\cite{parta2}   &TPAMI 2020   & {LiDAR}     & 87.81  & 78.49  & 73.51        &91.70  &87.79  &84.61    \\
 
         & STD~\cite{std}            &ICCV 2019   & {LiDAR}     & 87.95  & 79.71  & 75.09       &94.74 &89.19 &86.42   \\
         & 3D-CVF~\cite{3dcvf-net}           &ECCV 2020   & {LiDAR+RGB}   & 89.20  & 80.05  & 73.11      &93.52  &89.56 &82.45   \\
         & CLOCs PVCas~\cite{clocs-net}  &IROS 2020   & {LiDAR+RGB}     & 88.94  & 80.67  &77.15        &93.05  &89.80  &86.57  \\
         & PV-RCNN~\cite{pv-rcnn}          &CVPR 2020   & {LiDAR}     & 90.25  & 81.43  & 76.82        &94.98  &90.65  &86.14  \\
         & \emph{Voxel-RCNN~\cite{voxelrcnn}} &AAAI 2021 &{LiDAR}    & 90.90 & 81.62 &77.06   & 94.85 &88.83  & 86.13 \\
         \cline{2-10}
         & \bf \systemname (Ours)     & - & LiDAR+RGB   & 91.02 & \bf83.21 &\bf78.20      &93.02 &\bf91.86 &\bf86.94     \\

      \hline
   \end{tabular}
   \vspace*{1mm}
   \caption{Comparison with the state-of-the-art methods on the KITTI \textit{test} set for car detection, with 3D and BEV  precision of 40 sampling recall points evaluated on the KITTI server. Our \systemname achieves the highest precision for 3D moderate, BEV moderate and BEV hard. \systemname also ranks the 1st among all the published methods for 3D hard. }
   \vspace*{-3mm}
   \label{accuracy}
\end{table*}

\begin{table*}[t]
  \centering \addtolength{\tabcolsep}{-1pt}
  \footnotesize
  \begin{tabular}{|c|c|ccc|ccc|  }
      \hline
      Method &head & 3D Easy & 3D Moderate & 3D Hard    & BEV Easy   & BEV Moderate & BEV Hard     \\ 
      \hline\hline
      Voxel-RCNN~\cite{voxelrcnn} &LiDAR  &92.38 & 85.29 &82.86 &\bf95.52 &91.25 & 88.99 \\

      \systemname-base &LiDAR  & 93.15  & 83.58  & 80.04   & 94.30    & 91.72  & 89.30   \\ 
      \systemname  &Fusion  & \bf93.42   & \bf88.76 & \bf86.05 &94.11  & \bf92.44  & \bf89.88  \\
      \hline
  \end{tabular}\vspace{0.1cm}
  \caption{ Our results on the KITTI \textit{val} set for car detection. \systemname-base denotes the network when \systemname only takes the LiDAR point cloud information (with all the image features and the image only branch loss weight set to zero).  Voxel-RCNN is the point cloud backbone network that we have adopted. The results show that our \systemname achieves pronounced improvement compared with the baseline networks, especially for 3D metrics. }\vspace{-0.05in}
  \label{tab:kittival_compare}
\end{table*}

In this section, we discuss our experimentation details and evaluation results. We first present the datasets and metrics, as well as the implementation details. We then compare the performance of our approach against many recent methods on the test set. We also quantify the impact of the proposed network components by conducting relevant ablation studies. Furthermore, we conduct more experiments on validation set to perform qualitative analysis and inference time analysis  to demonstrate more salient properties of our methods.

\subsection{Dataset and Metrics}

The \textbf{KITTI dataset}~\cite{kitti-net} consists of 7,481 training frames and 7,518 testing frames, with 2D and 3D annotations of cars, pedestrians and cyclists on the streets. It consists of camera images and point clouds from a Velodyne HDL-64E LiDAR. Each class is further divided into three difficulty levels: easy, moderate and hard, according to their object size, occlusion level and truncation level. Note that among the most popular autonomous driving dataset KITTI,  nuScenes~\cite{nuScenes}, and Waymo~\cite{waymo}, only KITTI provides both LiDAR and stereo data. That is also the reason why we have only conducted experiments on KITTI. 

The \textbf{KINS dataset}~\cite{kins-net} was created by annotating amodal pixel-level annotation of instances to the KITTI dataset. The task aims to segment each instance  even under severe occlusion. It shares the same raw data with KITTI object detection dataset and the only difference is the additional annotation. KINS consists of 7,474 training samples and 7,517 testing samples.

The \textbf{metric} we use in the evaluation is mainly the 3D object detection Average Precision ($AP_{3D}$) metric. We consider a predicted box is true positive if its 3D intersection over union ratio (3D IoU) with a ground truth box is above a preset threshold. In our evaluation, the 3D IoU threshold for the \emph{car} category is 0.7.   
we only conduct evaluation on the `car' category because it contains more annotated boxes and gives more stable results than the other categories.   
Note that recently the KITTI benchmark has updated the evaluation protocol~\cite{evalprotocal}  and increased the recall positions from 11 to 40, aiming at a more fair evaluation result. All the evaluation experiments we have conducted follow this newest protocol.

\subsection{Implementation} 
 
\subsubsection{Training strategy}  
   The 7,481 frames on KITTI training set are further split into training set and validation set following the same setting as in~\cite{pointpillar,pv-rcnn,second-net,3dcvf-net}. The training set contains 3,712 frames that belong to 96 different scenarios. The validation set contains  3,769 frames from 45 different scenarios. In our experiments, We follow the same setting as in~\cite{pv-rcnn}; we project LiDAR  point clouds to the camera view, and preserve only the points located inside the view. When submitting to test server, we use all 7,474 training samples.

\subsubsection{Data Augmentation} 

Data augmentation is vital to preventing the networks from over-fitting.  We adopt four data augmentation strategies proposed in~\cite{pv-rcnn}. The fourth strategy is suitably modified as described in the design section. They are  (1) globally rotating the whole scene around the Z axis following the uniform distribution $ [-\frac{\pi}{2} , \frac{\pi}{2} ] $, (2) flipping the scene around the X axis with a probability of 0.5, and (3) scale transformation for each object following the uniform distribution of $ [ 0.95 , 1.05 ]$.  (4) ground truth boxes sampling. We generate the foreground object mask list  using the pretrained Deep Snake network~\cite{deepsnake}. The pretrained model was trained on the KINS training set following the default configuration. Then we associate the stereo-GT triplets, and randomly paste 15 triplets into the new scene. The 2D occlusion threshold $\tau_{2D}$ is set to 0.7 to simulate occlusion.  $\tau_{3D}$ is set to 0.  

\subsubsection{ Network setting}

We adopt Voxel-RCNN~\cite{voxelrcnn} as our 3D point cloud baseline. \systemname contains a 3D backbone,  a 2D RPN network, two voxel-RoI pooling layers and two detection heads. The 3D backbone takes raw points as input and divides them into voxels. The voxel resolution is $[.05 , .05 , .1]m$. Then voxels are processed by the 3D sparse convolution backbone as in~\cite{voxelrcnn}. Voxel features are then transferred to bird's eye view (BEV), and a 2D RPN network is applied to generate 3D region proposals. We enlarge each proposal for $0.8$m on each dimension. We follow most of the settings in Voxel-RCNN  with suitable modifications. Specifically, for each 3D scene, the range of LiDAR points are limited to $ [ 0 , 70 ]m $ for X axis, $ [ -3 , 1 ]m $ for Z axis,  $ [ -40 , 40 ]m $ for Y axis.We use the KITTI LiDAR coordinate system, so that Z axis is vertical to the earth.  We set the maximum points per voxel to be 5, the maximum number of voxels to be 40000.

As far as 2D image feature extraction is concerned, our 2D backbone adopts two 2D convolution blocks with the channel number $[32, 32]$ and a down sample rate $[1, 2]$. Each block contains two stacked 3 $\times$ 3 convolution layers.  Then we randomly rotate the 3D proposals from $[-0.08, 0.08]$. and random resize the rest dimensions from $[-0.15, 0.15]m$. Then we discretize proposals by sampling preset virtual points, the resolution of virtual points is 16 for width, 8 for length, and 22 for height. The virtual points are decorated by image features, and voxelized into 3D voxels. For feature aggregation, the resolution of query points $G_x = G_y = G_z = 6$.  The resolution of 3D voxel space is $[0.2 , 0.2, 0.1]$m.  When associating $D_0$, $D_1$, $D_2$  to virtual points and generating feature vectors,  following~\cite{voxelrcnn}, we take the Manhattan distance as the query range, and query in a multi-scale manner. For $Q_0$ the distance in X/Y/Z axis is set to $[2, 2, 2]$; for  $Q_1$ it is set to $[4, 4, 4]$.   We use the KITTI LiDAR coordinate system. 

During the training stage, we set the IoU threshold to 0.7, and sample 40 proposals as in~\cite{pv-rcnn,voxelrcnn}. During the evaluation stage, we set the IoU threshold to 0.1, and keep only 20 proposals. The score threshold is set to 0.1 during inference. Finally, the \systemname  network  was trained over 8 NVIDIA RTX 2080Ti GPUs by ADAM, with a batch size 16, learning rate 0.008, and weight decay of 0.01 for 80 epochs. It takes around 7.5 hours to complete the training for \systemname. 
 
\systemname was trained with a 3D voxel space resolution of  $[0.2, 0.2,0.1]$m, and  grid points resolution of  $16 \times 8 \times 22$. 
In \systemname, we generate 5K-30K non-empty voxels with a sparsity level less than 0.005. It has a similar sparsity level of typical LiDAR point clouds -- the point clouds in KITTI typically generate 5K-8K voxels with a sparsity of nearly 0.005 in the voxelization stage as stated in~\cite{voxelnet}.  

\subsubsection{Loss Function} 

We implement our loss function based on~\cite{pv-rcnn}. Suppose we use a vector $(x, y, z, w, l, h, \theta)$ to represent the ground truth box and anchors. Then we calculate the localization regression residuals accordingly between the ground truth box and the anchor $(\Delta x, \Delta y, \Delta z, \Delta w, \Delta l, \Delta h, \Delta\theta)$ by:
\[\Delta x = \frac{x_g-x_a}{d_a}, \Delta y = \frac{y_g-y_a}{d_a}, \Delta z = \frac{z_g-z_a}{d_a},\] 
\[\Delta w = log\frac{w_g}{w_a}, \Delta l = log\frac{l_g}{l_a}, \Delta h = log\frac{h_g}{h_a},\]
\[\Delta \theta = sin (\theta_g - \theta_a),\]
where $g$ is the ground truth box and $a$ is the anchor box, and we have $d_a = \sqrt{w_a^2 + l_a^2}$. 

We adopt the same region proposal loss $L_{rpn}$ as in~\cite{pointpillar}. Here focal loss\cite{lin2017focal} is utilized for anchor classification, and Smooth L1 loss is utilized for anchor box regression,
\[
L_{{rpn}} =  L_{{cls}} + \beta\sum_{{r} \in \{x, y, z, l, h, w, \theta\}} \mathcal{L}_{{smooth-L1}}(\widehat{\Delta {r}^{a}}, \Delta {r}^{a}), \]
where  $L_{cls}$  denotes  the anchor classification loss, $\widehat{\Delta {r}^{a}}$ denotes the predicted anchor box residual,  $\Delta {r}^{a}$ denotes the  target anchor box residual.

The combined multi-branch loss consists of three parts, the cross entropy IoU prediction loss, and two box regression losses from two branches:
\[
    L_{{rcnn}} =   L_{ {IoU}}  +   W_{ {V}}  \cdot L_{ {V}} +  W_{ {A }}  \cdot  L_{ {A}},\]
where $ L_{ {IoU}} $ is the cross entropy IoU prediction loss following~\cite{voxelrcnn,pv-rcnn}, and $L_{ {A}}$ and $L_{ {V}}$ are the Smooth L1 loss for 3D bounding box regression.  $ W_{ {V}} $ and  $ W_{ {A}} $ denote the proposed loss weights for the two branches. The regression target is the box residual following $L_{rpn}$.

\subsection{\systemname Detection Results on KITTI Test Set}
Table~\ref{accuracy} summarizes the 3D detection performance of \systemname  on the KITTI test set from the official online leaderboard as of July 14th, 2021. For the sake of space, we show published schemes here. The best results in each category are highlighted by the bold font. Among all the 27 networks, \systemname delivers the best results for the following 3 (out of 6) categories: $AP_{3D}$ moderate ( 83.08\%), $AP_{BEV}$ moderate ( 91.86\%), and $AP_{BEV}$ hard (86.94\%). Please note that \systemname delivers the best result on $AP_{3D}$ hard among all the published methods.  Please also note that our performance for the easy category is not the best, because objects in this category are near and have no/little occlusion. Thus the benefit of fusion is not very clear.  

In addition, compared with Voxel-RCNN, our point cloud backbone network, \systemname again achieves the performance gain on $AP_{3D}$ moderate (+ 1.59\%), $AP_{3D}$ hard (+ 1.14\%), $AP_{BEV}$ moderate (+ 3.03\%) and  $AP_{BEV}$ hard (+ 0.81\%).  These results clearly demonstrate that  \systemname is a powerful multi-modal 3D object detection network.

\subsection{\systemname Detection Results on KITTI Validation Set}  
Table~\ref{tab:kittival_compare} shows the $AP_{3D}$ and $AP_{BEV}$ results on the validation set. Here, we consider three networks: \systemname, Voxel-RCNN (our point cloud backbone), \systemname-base (with all the image features and image loss set to zero). 

We have the following observations. Firstly, \systemname performs the best in most of the cases, except for BEV Easy. This is because easy objects contain many LiDAR points that are already very accurate on BEV. Secondly, the performance gain on $AP_{3D}$ is more pronounced than that on $AP_{BEV}$. A possible reason is that image information contributes more to height estimation than other dimensions like rotation estimation.  Thirdly, \systemname-base fares worse than Voxel-RCNN. This is because the various multi-modal data augmentation techniques may possibly hurt the performance when the image features are not considered.

\subsection{Ablation Studies}
We have also conducted a set of ablation studies to evaluate the following important design choices. 

\vspace{3pt} \noindent\textbf{Impact of Network Components:} We first evaluate the impact of various network components.  Here, we remove different components from the overall \systemname  pipeline -- namely, virtual point image sampling, random 3D proposal resizing, right image, multi-branch loss -- and quantify the impact of these components one by one, as shown in Table~\ref{tab:ablation}.  Virtual point image sampling means the image features are sampled by virtual points instead of actual LiDAR points. We find that each of these four components contributes to the overall performance. Among the four, the virtual point image sampling  benefits $Mod_{3D}$  performance the most. It again proves the effectiveness of our fusion framework.    We noticed that stereo information and multi-branch loss  are more helpful for more  heavily occluded samples in  the hard category.  Also, since  multi-branch loss  emphasizes more on hard samples,  it may potentially hurt the   $Easy_{3D}$  performance.

\vspace{3pt} \noindent\textbf{Impact of Virtual Point Density and Sparse Convolution Layer Number:}
We also conduct experiments to study the impact of virtual point density as well as the impact of sparse convolution layer number (used for image feature extraction). Here, we down-sample the LiDAR beam number from 64 to 32 as these parameters bear little impact with 64-beam LiDAR. The results are shown in Table~\ref{tab:grid_res}. 
 We observe that decreasing the virtual point density hurts the performance, so does having fewer sparse convolution layers. 
We also observe that the  accuracy under the virtual point resolution of $[30 , 18,  30] $ is similar to the accuracy under  $[25, 12, 25 ]$. This is because $[25, 12, 25] $ is already a fine granularity. For example, for a typical car,  the vertical distance between two virtual points is around $10cm$ assuming a car is of 1.8m in height, and the proposals are enlarged by $0.8m$ in each dimension. 
	\begin{table}[b]
	\renewcommand\arraystretch{1}
	\small
 
	\begin{center}
		\scalebox{0.9}[0.9]{
			\setlength\tabcolsep{6pt}
			\begin{tabular}{|c|c|c|c|}
				\hline
				  method     & 3D Easy   & 3D Mod    & 3D Hard  \\   
				\hline
			      \systemname   & 93.21  & \bf 88.45  & \bf 85.76        \\  
			      w/o Virtual Point Image Sampling     &92.91   &85.49   & 82.91   \\  
			      w/o Random Resizing  & 92.37   &86.08   & 83.61    \\  
		 
			      w/o Right Image  &93.09  &88.17  &83.77                  \\   
				  w/o Multi-branch Loss     & \bf 93.38   &88.39   &83.79 \\  
  
				\hline
			\end{tabular}
		}

	\end{center}
	\vspace{0.2cm}
	\caption{ The ablation studies to evaluate the impact of our proposed modules }

	\label{tab:ablation}
\end{table}

	\begin{table}[b]
	\renewcommand\arraystretch{1}
	\small
 
	\begin{center}
		\scalebox{0.9}[0.9]{
			\setlength\tabcolsep{6pt}
	\begin{tabular}{|c|c|c|c|c|}
	
    				\hline
    				  method   &$\tau_{2D}$  & 3D Easy   & 3D Mod    & 3D Hard  \\ %
    				\hline
    			  w/o cut-n-paste    & -   &92.06   &85.83   & 83.51       \\  
    				 	\hline
   			      box cut-n-paste  & -  &92.85  &85.92 &83.25              \\  
   			        	\hline
    			    \multirow{4}*{ mask cut-n-paste(ours)  }  & 0.7  & 93.31  &  88.44  & 85.92      \\
                    \cline{2-5}
    			   &0.4   & 92.97  &  88.51  &  85.87    \\
    			                      \cline{2-5}
                   &0  &92.67  & 88.41  &83.91    \\
                                       \cline{2-5}
                   &1  &93.45  & 88.39  &83.98    \\ 

    				\hline
    			\end{tabular}
		}

	\end{center}
	\vspace{0.2cm}
	\caption{    Effectiveness of the cut-n-paste augmentation module    } 
	\label{tab:aug}
\end{table}

\begin{figure}
\centering
 
\includegraphics[width=0.9\linewidth]{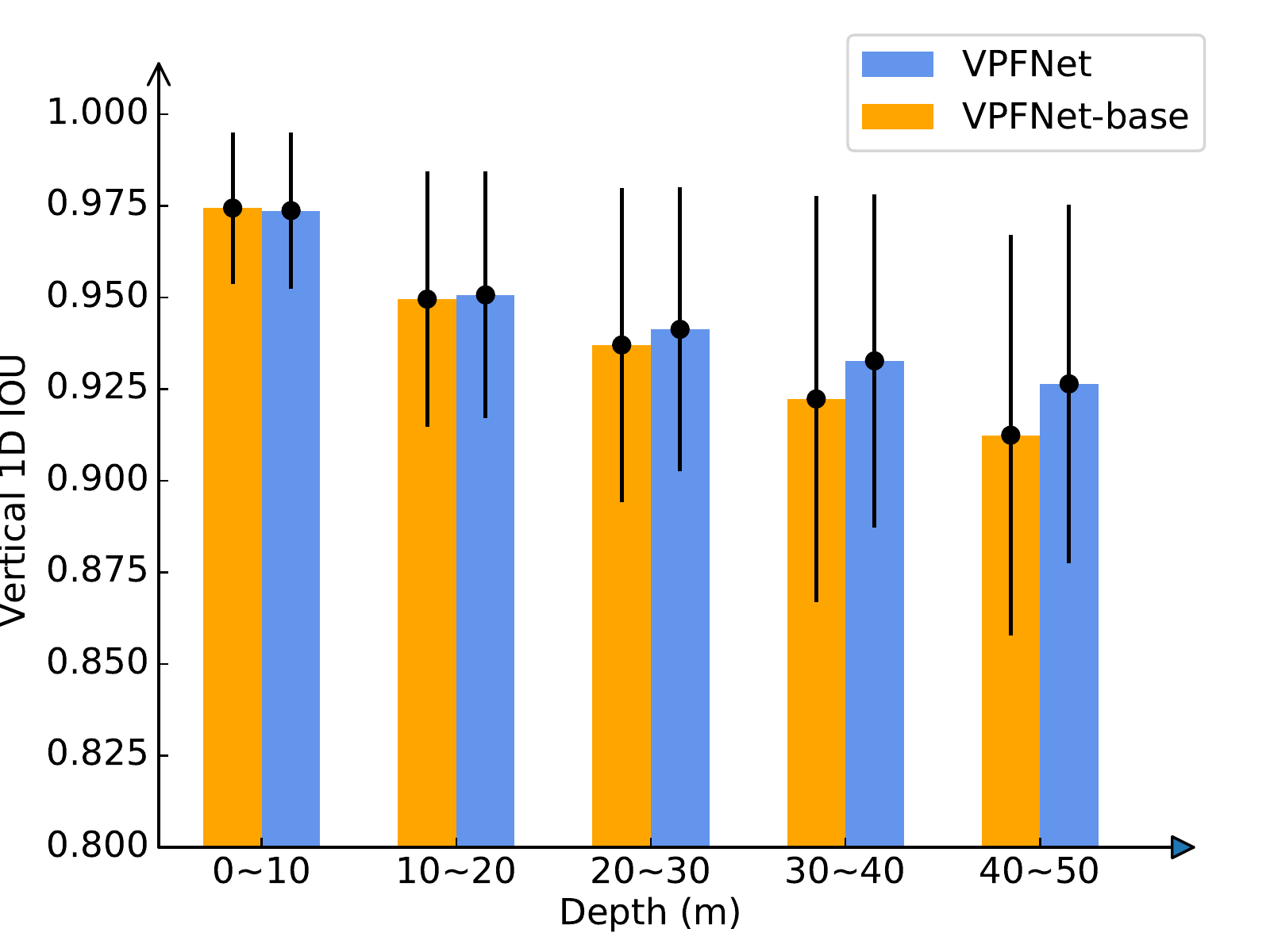}
 
\centering
\caption{  \label{fig:distance} \systemname can reduce height estimation errors compared to \systemname-base in all depth ranges, especially for faraway objects. Y-axis is the mean value and variance of 1D IOU between GT 2D boxes and  predicted 2D boxes  }
\end{figure}

\vspace{3pt} \noindent\textbf{Better Height Estimation:}   We have made the hypothesis that \systemname can effectively improve the height estimation compared to \systemname-base, due to the help of image features. We show the results in Figure~\ref{fig:distance}. The X-axis is the depth value,    and the  
 Y-axis is the mean value and variance of 1D IOU between GT 2D boxes and  predicted 2D boxes.  The 1D IOU is calculated  along the vertical dimension  of  2D boxes. We observe that the height estimation is improved  in all  depth ranges, especially when depth is between 30-50 meters.

\vspace{3pt} \noindent\textbf{Impact of Data Augmentation:}  In Table~\ref{tab:aug}, we compare the detection results when we have no cut-n-paste, box cut-n-paste as in~\cite{augmenting}, as well as our mask cut-n-paste with different $\tau_{2D}$ values. The results show that our mask cut-n-paste augmentation leads to an improvement of 1.25\%,  2.61\%, and 2.41\% for  $Easy_{3D}$,  $Mod_{3D}$, and  $Hard_{3D}$, respectively. Meanwhile, according to Table~\ref{tab:ablation} in our paper, the performance improvement brought by our model design is 0.3\%,  2.86\%, and 2.85\% for  $Easy_{3D}$,  $Mod_{3D}$, and  $Hard_{3D}$,  respectively. Comparing the two, our model design yields larger improvements than our augmentation scheme on  $Mod_{3D}$ and  $Hard_{3D}$,.

\begin{table}[b]
	\centering 
	\footnotesize
	\setlength\tabcolsep{3pt}
	\begin{tabular}{|c|c|c|c|c|c|c| }
		\hline
		$N_x$ &$N_y$ &$N_z$ &$N_{layer}$   & time(ms) &  3D Hard   &  BEV Hard    \\
    	\hline                        
		 30  & 18  & 30    & 6  &102  &\bf83.25  & 89.35  \\
    	\hline                        
		25 &12 &25    & 6  &81  & 83.24  &\bf89.43  \\
		\hline
		25 &12 &25    & 1  &61  &81.20  &89.08  \\
		\hline
	    16 &8 &25    & 6  &76    &81.11 &89.37  \\
		\hline
	    	 16  & 8  & 22     & 6  &71    &81.16 &89.12  \\
		\hline
		16 &8 &16    & 6  &62    &81.03 &87.30  \\
		\hline
		
	\end{tabular}
	\vspace{0.2cm}
	\caption{  $AP_{3D}$ and inference time   with several virtual point density settings (denoted by $N_x, N_y, Nz$) and  sparse convolution layer number (denoted by $N_{layer}$) in \systemname.  Note that the experiments were conducted on the down-sampled KITTI data, with LiDAR down-sampled to 32 beams. }
	\label{tab:grid_res}
 
\end{table}

\begin{figure*}
\centering

\subfigure[road scenes]{
\includegraphics[width=0.45\linewidth]{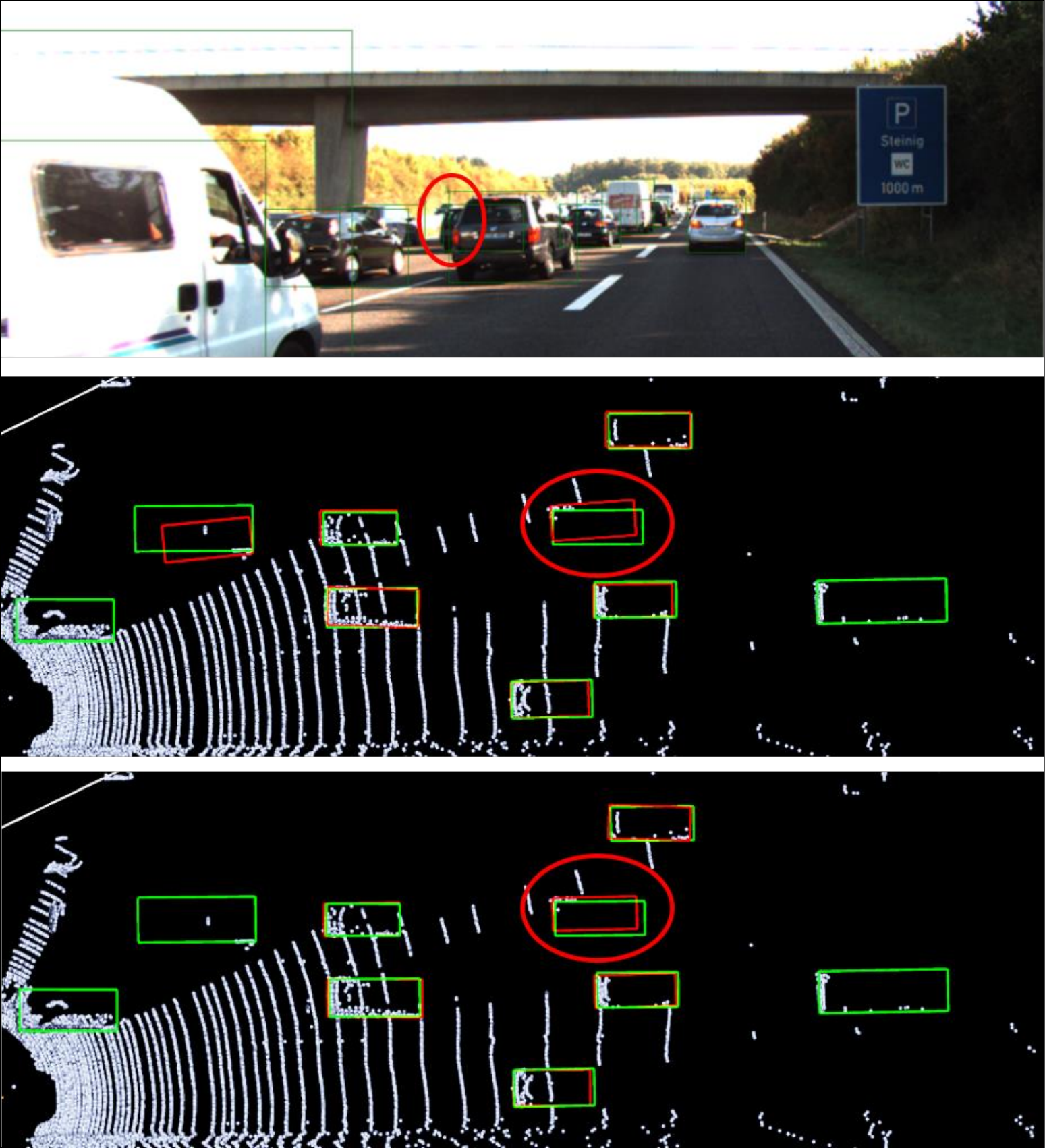}
}%
\subfigure[residential scenes]{
 
\includegraphics[width=0.47\linewidth]{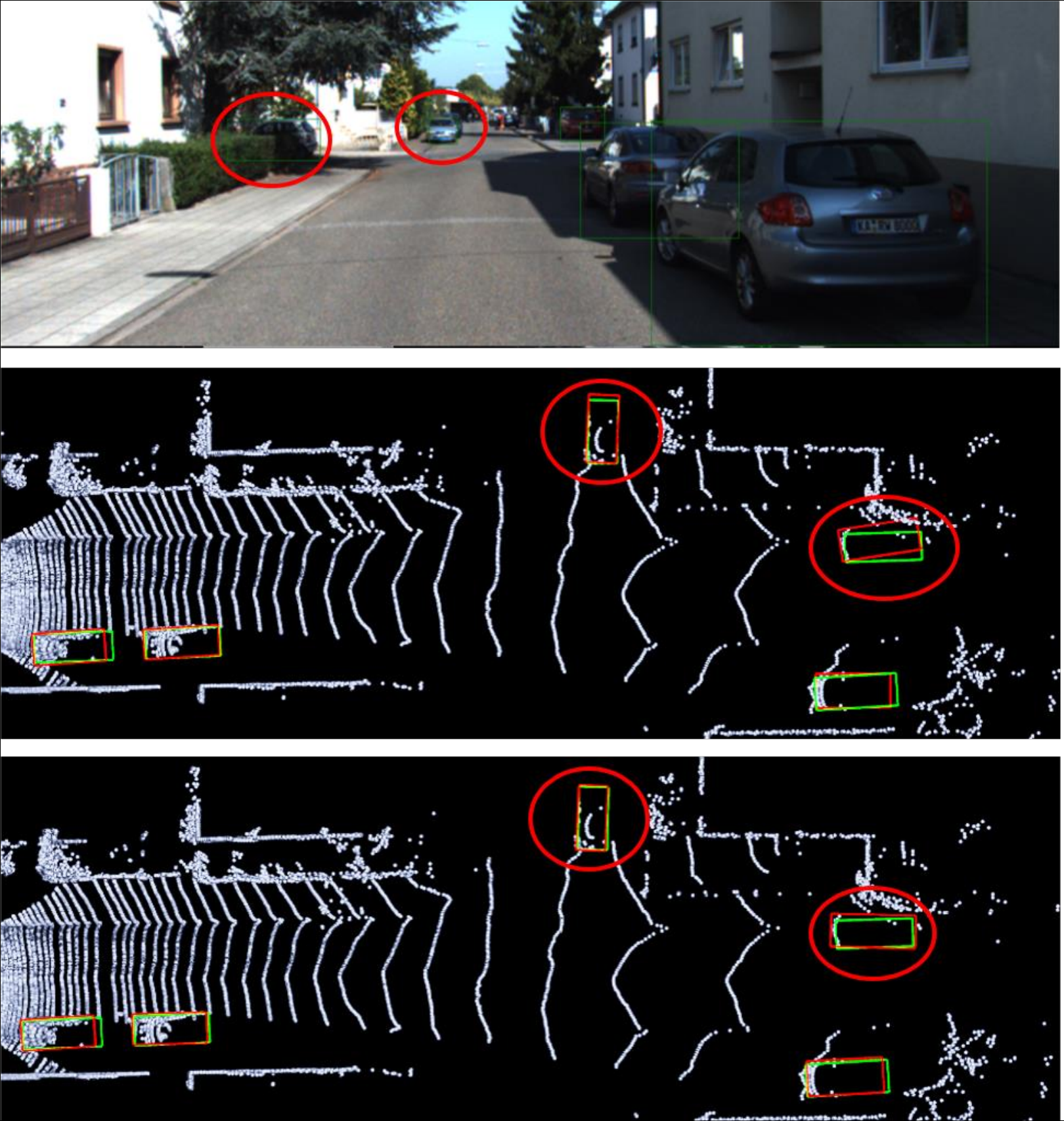}
}%

\centering
\caption{\label{fig:quanti} Qualitative results for (a) road scenes and (b) residential scenes. From top to bottom, we have the left image, detection on bird's eye view from \systemname-base, and detection on bird's eye view (BEV) from our proposed \systemname. Green boxes on the picture mark ground truth boxes while red boxes mark prediction boxes. Predictions with scores $> 0.1$ are visualized on the pictures. We highlight some areas in red circle where the BEV detection results are improved significantly. }
\end{figure*}

\begin{figure*}[!t]
\centering

\subfigure[ Results for a far-way object. Image features are activated on the top and the wheels of the car. ]{
 
\includegraphics[width=1\linewidth]{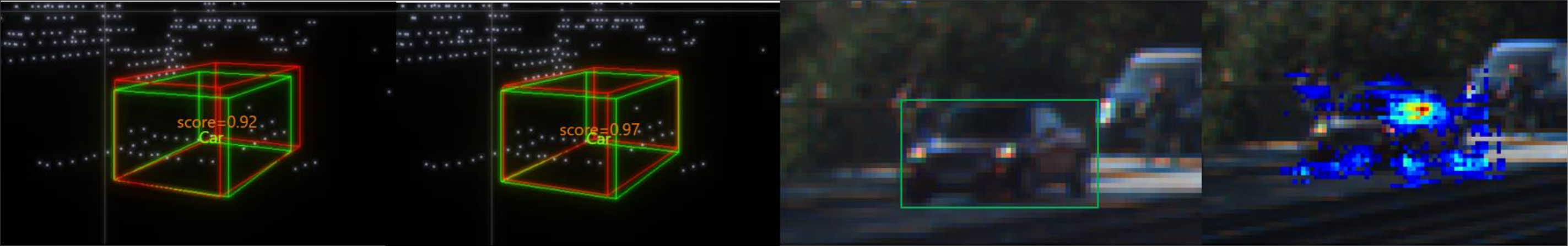} 
} \\
\vspace{2pt}

\subfigure[ Results for an occluded object.  Image features are activated  on the bottom of the occluded  car.  ]{
\includegraphics[width=1\linewidth]{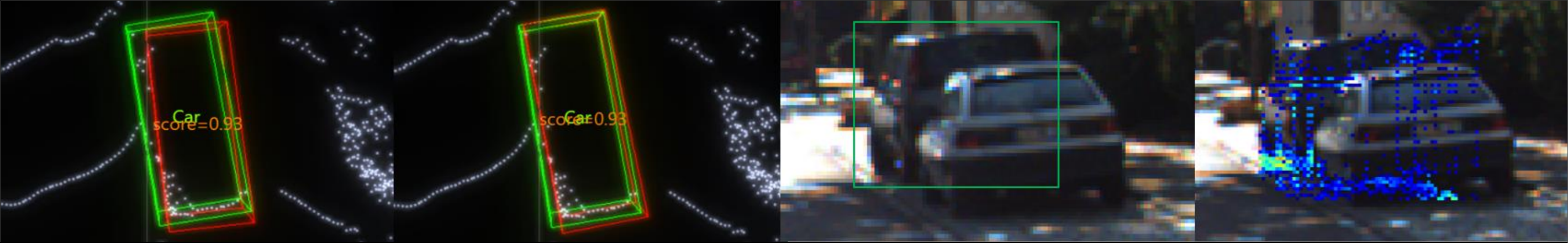} }
\\
  
\vspace{2pt}

\subfigure[ Results for a near object. The box length are improved. We can observe that our sample density is lower than that of the image pixels.  ]{
\includegraphics[width=1\linewidth]{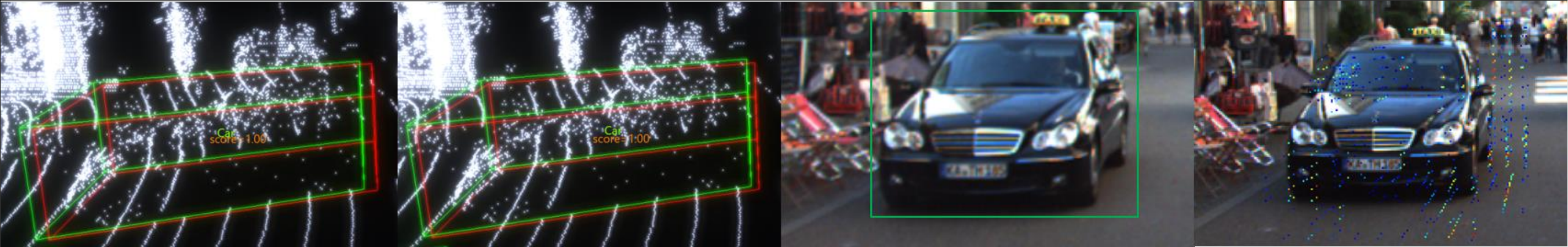} }

\vspace{2pt}
 \caption{\label{fig:quanti2} Qualitative results. From left to right: detected object from \systemname-base, detected object from our \systemname, the object in left image, the visualization of  the  intermediate feature vectors ($D_0$).  Green boxes on the picture mark ground truth boxes while red boxes mark prediction boxes. Predictions with scores $> 0.1$ are visualized on the pictures.  We show that \systemname can improve both box size and rotation estimation on near, far, and occluded situations. }
\end{figure*}

\subsection{Qualitative Results} We also include some quantitative results in Figure~\ref{fig:quanti} and Figure~\ref{fig:quanti2}, where we visualize the results from \systemname and \systemname-base. Figures~\ref{fig:quanti}(a)-(b) show that though the box predictions  (marked in red) generated by \systemname-base already have high IoU with the ground truth boxes (marked in green), \systemname can still improve the accuracy on the bird's eye view (BEV). Figure~\ref{fig:quanti}(a) shows an example in which the rotation estimation of an occluded car is improved. Figure~\ref{fig:quanti}(b) shows the rotation estimation of a far-away object and the box size estimation of an occluded box are improved.

Figures~\ref{fig:quanti2} (a)-(c) show  the results from both detectors, as well as  the  intermediate feature vectors ($D_0$) after sparse convolution. In Figure~\ref{fig:quanti2}(a), the predicted box height from \systemname-base  exceeds the actual size because LiDAR beams locate only on the middle part of the car. The activated virtual point features are mainly located on the top of the car and the four wheels.  In  Figure~\ref{fig:quanti2}(b), the predicted orientation from  \systemname-base  shows a drift because the object suffers from a severe occlusion. Figure~\ref{fig:quanti2}(c) shows that even for near objects, the detection result can be slightly improved by \systemname.

\subsection{Inference Time Measurement}

In our design, the inference time is mainly affected by the density of virtual points and query points. The submitted version of \systemname partitions each proposal into $16 \times 8 \times 22$ grid blocks and  $ 6 \times 6 \times 6$ query points, resulting in  inference time of 63.6 ms and a FPS of 15.7 on a single NVIDIA RTX 2080Ti GPU. Note that there is less than 140 actual LiDAR points located inside of each 3D proposals on average. Table~\ref{tab:grid_res} shows the overall inference time of \systemname with different grid block numbers. We observe that although the accuracy increases with the density of virtue points, the latency increases even faster. Therefore it is important to find a trade off between inference time and accuracy.

\section{Conclusion}

In this work, we present virtual-point fusion  to combine LiDAR and stereo data for more accurate 3D object detection. The key difference from existing methods lies in the fact that we take into consideration the resolution mismatch between the two sensors.  We employ virtual points whose density is in between that of 3D points and 2D pixels to bridge the resolution gap and to carefully balance data sampling rate and computing efficiency. With virtual points, we can efficiently sample image features while aggregating $K$ nearby point cloud features. Further, final predictions are obtained by having two detection heads with different weights to prevent single-modality dominance. Finally, multi-modal data augmentation techniques are devised to further boost the performance. Experiments on the KITTI dataset prove the effectiveness of the proposed method, which performs the best for several KITTI metrics and supports an FPS of 15.7.

\ifCLASSOPTIONcaptionsoff
  \newpage
\fi

\bibliographystyle{IEEEtran}

\bibliography{ref}
\end{document}